\documentclass[sigconf]{acmart}
\renewcommand\footnotetextcopyrightpermission[1]{} 
\usepackage{amsmath}

\usepackage{amssymb}
\usepackage{algorithm}
\usepackage{algorithmic}
\usepackage{caption}
\usepackage{booktabs}
\usepackage{graphicx}
\usepackage{makecell}
\usepackage[export]{adjustbox}

\AtBeginDocument{%
  }

\begin{document}

\title{Defending Against Backdoor Attack on Graph Nerual Network by
Explainability}

\author{Bingchen Jiang}
\affiliation{%
  \institution{Central South University}
  \country{China}
}

\author{Zhao Li}
\affiliation{%
  \institution{Central South University}
  \country{China}}

\begin{abstract}
			Backdoor attack is a powerful attack algorithm to deep learning model. Recently, GNN's vulnerability to backdoor attack has been proved especially on graph classification task. In this paper, we propose the first backdoor detection and defense method on GNN. Most backdoor attack depends on injecting small but influential trigger to the clean sample. For graph data, current backdoor attack focus on manipulating the graph structure to inject the trigger. We find that there are apparent differences between benign samples and malicious samples in some explanatory evaluation metrics, such as fidelity and infidelity. After identifying the malicious sample, the explainability of the GNN model can help us capture the most significant subgraph which is probably the trigger in a trojan graph. We use various dataset and different attack settings to prove the effectiveness of our defense method. The attack success rate all turns out to decrease considerably.

\end{abstract}


\maketitle

\section{Introduction}
    Machine learning (ML) models are increasingly deployed to make decisions on our behalf on various (mission-critical) tasks. In the fields of computer image processing, natural language processing and complex applications, medical image analysis, automatic driving, face payment, etc. have achieved very remarkable results.

    However, the neural network (referred to as: model) with deep learning as the core has some shortcomings. For example, the function of the model can be easily disabled after being attacked by malicious attack such as backdoor attack\cite{ref10,ref34}. Specifically, in backdoor attack:

     \begin{enumerate}
        \item  The backdoor is embedded so stealthily that it cannot be easily discovered, but an attacker can activate the backdoor using specific means to cause harm. For example, adding predefined conditions ("triggers") to the training data. During the model's inference phase, all the trigger-embedded samples are misclassified to the attacker's pre-defined target label.(i.e. attack effectiveness) 
        \item  Backdoor attack will not affect the normal performance of the system. In a deep learning system, that is, it will not affect or significantly reduce the prediction accuracy of the model for clean samples(i.e. attack evasiveness). 
     \end{enumerate}

    Backdoor detection and defense have been extensively studied in the image domain \cite{ref2,ref3,ref4,ref5}.With the development of GNN, researchers have also found that GNN can be easily attacked by backdoor attack \cite{ref26,ref27,ref78}. In order to implement backdoor attack in graph data, attackers must find a new kind of trigger. Unlike images whose data is structured and continuous, graph data is inherently unstructured and discrete, requiring triggers to be of the same nature. In the graph classification task, GNN predicts according to the topological structure and node features. It’s easy to come up with an idea that attackers can use a subgraph as a trigger. After injecting the triggers into training data, the model will acquire a “shortcut” function from the trigger structure to the attacker’s target label. During the model's inference phase, if the input graph includes the pre-defined trigger structure, GNN model will predict the target label. Until now, backdoor detection and defense on GNNs are unexplored. Concretely, it is because backdoor defense on graph data has the following difficulties.

    \subsection{Defense is Challenging}

    \begin{itemize}
    \item[$\bullet$] For those malicious image training sample which is marked by a wrong label, we can easily detect them by our eyes. For example, it's easy for us to identify the label of the picture in most dataset such as MINIST dataset and CIFAR-10 dataset. But for graph data, most of the time we can't discriminate the true label only via observing the topological structure and node features.

    \item[$\bullet$] Triggers can be of any shape and pattern, may be located anywhere on the input, and have indeterminate size. Moreover, the trigger can be a sparse subgraph or dense subgraph. It is extremely hard to predict the properties of the attacker's trigger. Consequently, some methods like dense-subgraph detection is invalid. In other word, defenders must prepare to cope variety of triggers.
    \end{itemize}

     \subsection{Our work and contribution}
     Given these challenges, we propose the first backdoor detection and defense method for GNNs. We reveal that triggers are much more influential than benign subgraphs. Compared to clean samples, trigger-embedded samples show completely different characters in some explainability evaluation metrics such as fidelity and infidelity. In essence, removing the trigger in the graph will affect the prediction probability considerably, whereas removing a benign subgraph will not cause the such condition.  In this paper, we introduce Explainability score(ES) to quantify such characteristics and to distinguish the malicious samples and clean samples. 

    After identifying the backdoor samples, we utilize the explainability method to find the most influential subgraph in backdoor samples, that is, the trigger. Once we delete the trigger, this malicious sample will transform into a clean sample. Then this backdoor attack is not effective anymore.

    We summarize our contributions as below:
    
    \begin{enumerate}
        \item We perform the first backdoor detection and defense method on GNNs. Our approach detects whether the input is trojaned or not. Once we find a malicious input sample, our approach can delete the trigger in this graph. Hence invalidate the backdoor attack.
        \item Our method can cope with a variety of triggers. That is to say, no matter what structure the trigger is, our method still performs well.
        \item We evaluate our method on three popular graph classification datasets with different attack settings. We achieve both low false acceptance rate(FAR) and low false rejection rate(FRR).
    \end{enumerate}

        \begin{figure*}[htbp]
        \begin{center}
        \includegraphics[width=1.0\textwidth]{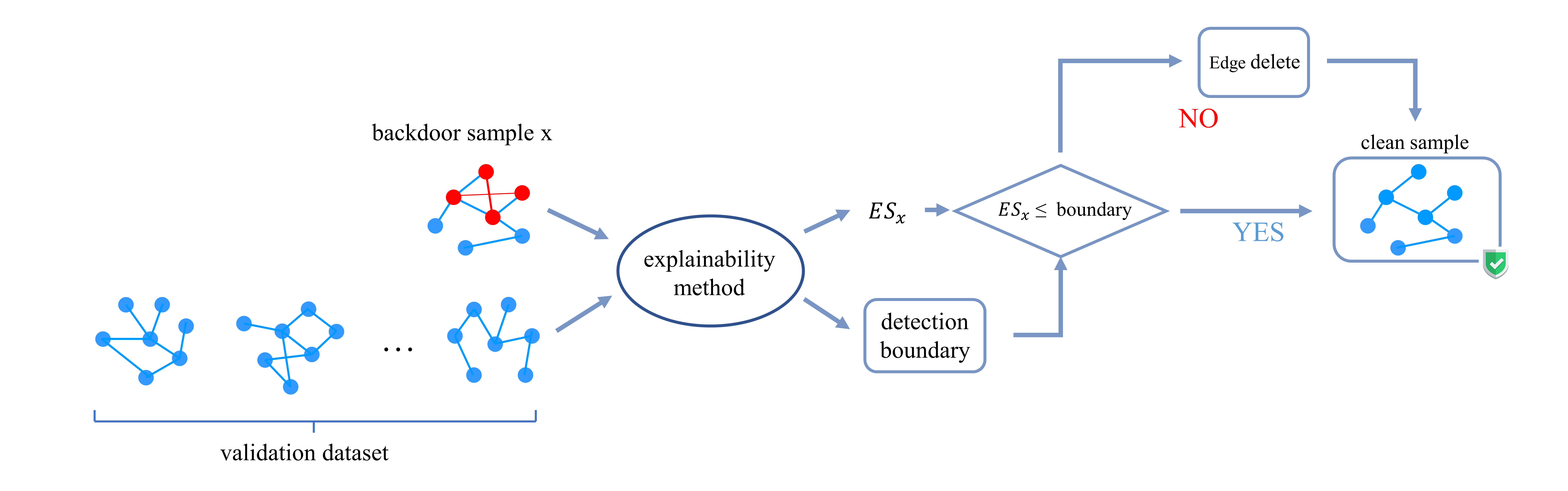}
        \end{center}
        \caption{backdoor detection and defense system}
        \label{fig:defense}
        \end{figure*}

\section{Background}
	\subsection{Graph Neural Network}
	
	\textbf{Graphs} have been widely used to model complex interactions between entities. For instance, In an urban transportation network, cities and connected roads can be modeled as a graph, where cities are nodes, and edges between two nodes represent road connections between cities. In a social network, a user and his friends can be modeled as a graph, where the user and his online friends are nodes, and the edges between the two nodes represent the relationships and interactions between them. 
	
	\noindent\textbf{Graph Neural Network (GNN)} takes a graph as an input, including topological structures and node features, and generates a representation $z_v$ for each node $v$. Most GNNs work with a neighborhood aggregation strategy:
	\begin{equation}
	    Z^{(k)} = AGGREGATE(A,Z^{(k-1)};\theta^{(k)})
	\end{equation}
	where $A$ represents for the adjacency matrix of the input graph and $Z^{(k)}$ is the node embeddings after k-th iteration. It also depend on the trainable parameters $\theta^{(k)}$ and the node embeddings $Z^{(k-1)}$ from the previous iteration. $Z^{(0)}$ is usually initialized as input graph's node features. In order to obtain the graph embeddings $z_G$, the READOUT functions pools the node embeddings $Z^{(k)}$ from the final iteration $k$:
    \begin{equation}
        z_G = READOUT(Z^{(k)})
    \end{equation}

    \noindent\textbf{Graph classification} is a basic graph analytics tool and has many applications such as, in cheminformatics, the mutagenicity, toxicity, anticancer activity, etc. of compound molecules are judged by classifying molecular graphs\cite{ref1,ref7}; In bioinformatics, protein network classification is used to judge whether a protein is an enzyme or not. It has the ability to treat a certain disease \cite{ref8,ref9}. From this point of view, graph classification research is of great significance. GNN based graph classification extends neural network to graph data. We can build an end to end framework for graph classification by applying a multilayer perceptron and a Softmax layer to graph representations.
	
    \subsection{Attack method}
	\textbf{adversarial attack:}The goal of the adversarial attack is to select a class in advance and then choose the attack method. Make the model classify the sample into another class as much as possible, and increase the difference between the $class_{new}$ probability and the original class probability. The adversarial attack happens in the model's inference phase. There are basically two kinds of adversarial attack: targeted attack and untargeted attack. The difficulty of these two kinds of adversarial attack are different. In targeted attack, the attacker wants the model to generate a specific wrong label. In untargeted attack, attacker doesn't focus on making the model wrongly generate a specific label. As long as it is not divided into the correct label. One of the challenges of adversarial attacks is that the change is not obvious. The measurement standard of the image field can be that the changed image looks similar to the eyes. In the graph field, we can ensure unnoticeable perturbations in our setting through graph structure preserving-perturbations and feature statistics preserving perturbations.\cite{ref22,ref24,ref23}.

     ~\\

        \noindent\textbf{backdoor attack:}Compared with adversarial attack, backdoor attack happens throughout the entire process. During the training phase, attacker hope to inject malicious trigger into target systems in some way. The injected backdoor is triggered by the attacker. During the inference process, if the input sample doesn't contain trigger, the backdoor will not be activated and the attacked model behaves similarly to the normal model temporarily; But if a trigger-embedded sample is fed into the model, then the backdoor will be activated and the output of the model becomes the target label pre\-specified by the attacker to achieve malicious purposes.
        
        Currently, poisoning training data is the most direct and common method of backdoor attacks. In poisoning-based attacks, the attacker modifies some training samples through pre\-set triggers (such as a small subgraph). The labels of these modified samples are replaced by the target labels specified by the attacker, resulting in poisoned samples\cite{ref25,ref26,ref27,ref1}. These poisoned samples and benign samples will be used for training at the same time to obtain a model with a backdoor.

    \subsection{Research about the explainability in GNN}
	In recent years, several methods for Explainability in Graph Neural Networks have been proposed, such as XGNN \cite{ref41} , gnnexplainer \cite{ref42}, PGExplainer \cite{ref43} , etc.These approaches focus on different aspects of graphical models and provide different perspectives to understand these models. They generally start from several questions to realize the interpretation of the graphical model: Which input edges are more important? Which input nodes are more important? Which node features are more important?
    
    Explanation techniques of deep learning models aim to find the parts of the input data which is the most influentialto the prediction results, providing input-dependent explanations for each input graph. And depending on how feature importance scores are obtained, these approaches can be divided into four distinct branches:

    \textbf{Gradient/feature-based methods} \cite{ref49,ref50} employ gradients or eigenvalues to represent the importance of different input features.
    \textbf{Perturbation-based methods} \cite{ref42,ref43,ref49,ref52,ref44} monitor the changes in predicted values under different input perturbations, thereby learning the importance scores of input features.
    \textbf{Decomposition-based methods} \cite{ref49,ref50,ref54,ref55} first decompose prediction scores, such as predicted probabilities, to the neurons of the last hidden layer. Such scores are then back-propagated layer by layer until the input space, and the decomposition score is taken as the importance score.
    \textbf{Agent-based methods} \cite{ref56,ref57,ref58} first draw samples from a dataset from the neighbors of a given example. The next step is to fit a simple and interpretable model, such as a decision tree, to the sampled data set. Interpretation of the original predictions is achieved through the interpretive surrogate model.

    \subsection{Threat Model}
        Our threat model is largely inspired by backdoor attacks in the image domain \cite{ref6,ref10,ref30,ref34,ref38,ref39}.
	\subsubsection{Attakers' goal and capability}
 ~\\
        \textbf{Attacker’s goal:}An attacker has two goals. First, the backdoored model performs well with most benign inputs, which makes the backdoor attack stealthy(attack evasiveness). Second, when the input contained attacker-predefined triggers, it showed targeted misclassification(attack effectiveness).
        
        \textbf{Attacker’s capability:}The attacker can poison a small fraction of training graphs in the training dataset. Specifically, the attacker can inject a trigger to some training graph, and change these samples' labels to an attacker-chosen target label. For instance, malicious users under an attacker’s control can provide the user with poisoned training graphs. Moreover, the attacker can inject the triggers into testing graphs.

	\subsubsection{Defenders' goal and capability}
~\\
        \textbf{Defenders' goal:}A defender has three goals. 
        First, the defender needs to identify the backdoor samples as much as possible.
        Second, the defense method should not influence the accuracy on benign samples.
        Finally, the defender disables the backdoors. For instance, the defender can remove malicious samples or patch the network to remove backdoors).

        \textbf{Defenders' capability:}From the defender side, we reason that defenders hold a small collection of validation datasets which is used to set the detection boundary between the clean sample and malicious sample.

\section{Our Explainability based Backdoor Defense}
    
    \subsection{Defense System Design}
	Our defense system is dipicted in Figure \ref{fig:defense} and summarized in Algorithm 1. There are mainly two processes in our defense system, namely backdoor detection and backdoor defense. The key to our backdoor detection is to identify the malicious samples successfully. 
    
    How to search and remove the triggers is the key to our backdoor defense. For those suspected samples, we manage to delete the trigger in it to invalidate the backdoor.  After removing the trigger, this backdoor sample can be regarded as a clean sample so that we can feed into the model again to get the original output.
    
    \textbf{backdoor detection:}Defender holds a small collection of clean samples called validation dataset. In the validation dataset, we calculate the explainability score(ES) for each graph. In particular, ES will be detailed in Section 3.2.4. We use the highest ES in validation dataset as a detection boundary to facilitate the classification on whether the input sample is malicious or not.

    Then we can deploy our defense system. We calculate the ES for every incoming input x.  If the ES is higher than the detection boundary, it is judged as a malicious sample, and if it is lower than the detection boundary, it is judged as a benign sample.

    \textbf{backdoor defense:}  Once we suspect an input graph is malicious. Since the attack's effectiveness depends on the trigger, there is no denying that the trigger must be the most influential subgraph in this graph. Using the explainability of the model can help us find the most important subgraph. That's just what we want to delete.

	\begin{algorithm}
	 	\renewcommand{\algorithmicrequire}{\textbf{Input:}}
	 	\renewcommand{\algorithmicensure}{\textbf{Output:}}
	 	\caption{Backdoor detection and defense Algorithm}
	 	\label{alg1}
	 	\begin{algorithmic}[1]
              \REQUIRE validation dataset $S$, $GNN_\theta()$,input sample $x$
              \ENSURE clean sample
                    
	 	 \STATE \textbf{For}  graph $i$ \textbf{in} $S$
	 	 \STATE \ \ \ \ Calculate $ES$ for graph i
	 	 \STATE Choose the highest $ES$ in validation dataset as the $detection$ $boundary$
	 	 \STATE Calculate $ES$ for $x$ 
              \STATE \textbf{If} $ES_x \geq$ $detection\ boundary$

              \STATE \ \ \ \ $explanation$ = explain($x$,$GNN_\theta()$)
              \STATE \ \ \ \ x = delete edge($explanation$, x)
              \STATE Return x
	 	\end{algorithmic}  
	 \end{algorithm}

                    


    \subsection{Explainability in GNN}
        Intuitively, as defenders, we do not know the trigger's information. Although the triggers can be of a variety of patterns and structures, they still have one thing in common -- trigger always plays an extremely important role in the model's prediction. Then the problem that how to discriminate between malicious samples and benign samples becomes the problem that how to distinguish the importance of trigger subgraph and benign subgraph. The Explainability of the model can help us identify input features that are important to the results. At the same time, there are two evaluation metrics called fidelity and infidelity which can quantify the importance of the features we find. This is the inspiration for our solution. 

        First, the explanation should be faithful to the model. To evaluate this, the Fidelity \cite{ref50} metric was recently proposed. The key idea is that if the important input features (node/edge/node features) identified by the explainability technique are important to the classification result, then the model's predictions should change significantly when those features are removed. As for the benign samples, there is always a benign subgraph which can increase the probability that the model will make a correct prediction. But for the backdoor samples, with the appearance of the trigger, increases the likelihood that the model will predict the attacker's pre-defined target label. Moreover, this increment is much bigger than the former, otherwise, our model will still make a correct prediction rather than the target label. In other words, as long as we try to remove the most important features in the sample, the probability of the original prediction will reduce. But for the backdoor sample, this reduction will be much bigger than the benign sample. 

        Concretely, In order to distinguish the importance of different edges to the classification results. explainability technique generates the importance map $m_i$. The explanations can be considered as a hard importance map $m_i$ where each element is 0 or 1 to indicate if the corresponding feature is influential. If the graph contains 37 edges. Then the explainability method can generate a (37,) tensor called edge mask. Each element indicates the importance of this edge for our model to make this prediction. For some methods like GNNExplainer, the important values are continuous values from 0 to 1. Then the importance map $m_i$ can be obtained by normalization and thresholding.

        \subsubsection{Fidelity}, defined as the difference in probability between the original prediction and the new prediction after removing important input features\cite{ref77}, which is a measure of the difference between the two prediction results. Several techniques are proposed to generated the importance map \cite{ref42,ref44,ref49,ref59}.Finally, the Fidelity of prediction accuracy can be computed as:

        \begin{equation}
                \text { Fidelity }=\frac{1}{N} \sum_{i=1}^{N}\left(f(\mathcal{G} i)_{y_{i}}-f\left(\mathcal{G} i^{1-m i}\right)_{y_i}\right)
        \end{equation}

        where $y_i$ is the original prediction of graph $i$ and $N$ is the number of graphs. Here $1-m_i$ means the complementary mask that removes the important input features. Higher fidelity indicates better explanations
        results and more discriminative features are identified. For malicious samples, trigger is its "only creed", and deleting most triggers will lead to judgment results change, so the Fidelity value of backdoor samples will be much larger than the clean sample.

        \subsubsection{Infidelity}, In contrast to the Fidelity metric, which studies prediction changes by removing 
important node/edge/node features, The Infidelity metric investigates prediction changes by retaining important input features and removing unimportant features. Obviously, important features should contain discriminative information, so even if unimportant features are removed, they should lead to similar predictions to the original predictions. Formally, the metric Infidelity can be calculated as:

        \begin{equation}
                \text { Infidelity }=\frac{1}{N} \sum_{i=1}^{N}\left(f(\mathcal{G} i)_{y_{i}}-f\left(\mathcal{G} i^{m i}\right)_{y_i}\right)
        \end{equation}

        where $\mathcal{G} i^{m i}$ is the new graph when keeping important features of $G_i$ based on explanation $m_i$. Unlike fidelity, lower infidelity values indicate less importance information for deletions and therefore better explanations of results. Compared with benign samples, the unimportant features in backdoor samples seem more unimportant. Backdoor samples tend to have lower infidelity.

        Now we own the method to distinguish between backdoor samples and benign samples. In order to invalidate backdoor, we only need to remove the most important edges in backdoor samples according to explanations. But there is still one critical problem. Because many advanced explainability method output continuous important values, how to set a proper threshold between important edges and unimportant edges. Concretely, for backdoor samples with big trigger size, deleting less edges is not enough to remove the whole trigger. The trigger may still work. For backdoor samples with small trigger size, although deleting edges can easily remove the trigger, it can also do damage to the natural structure of the graph because we may delete some edges belonging to benign subgraphs.

        In essence, we need to estimate the trigger size approximately. But actually, the defender doesn't know anything about the structure of the trigger. Hence we need to find other things that are relevant to the trigger size. Trigger size determines the number of edges in the trigger. That is to say, trigger with a big trigger size owns more edges. In that case, there will be a lot of really important edges. Compared with small size trigger, the importance map $m_i$ of big size trigger is more concentrative. 
        
        For example, for a malicious sample with 100 edges, we assume the trigger has 80 edges. It's pretty hard to say which edges are the most important. Simply speaking, these 80 edges form the trigger together. The influence of the trigger is distributed to 80 edges. Consequently, there is no big difference in importance among these edges. But if the trigger has 10 edges. It's apparent that these 10 edges will be extremely important compared with other 90 edges. It's just like the situation that there are 100 persons. Distributing 100 dollars to 80 persons will not cause a huge gap of wealth among them. While distributing 100 dollars to 10 persons will increase the gap of wealth to some degree. To sum up, small size trigger's importance map $m_i$ will have a high degree of dispersion while big size trigger will not.

\begin{figure}[h]
		\centering
		\includegraphics[width=0.8\linewidth]{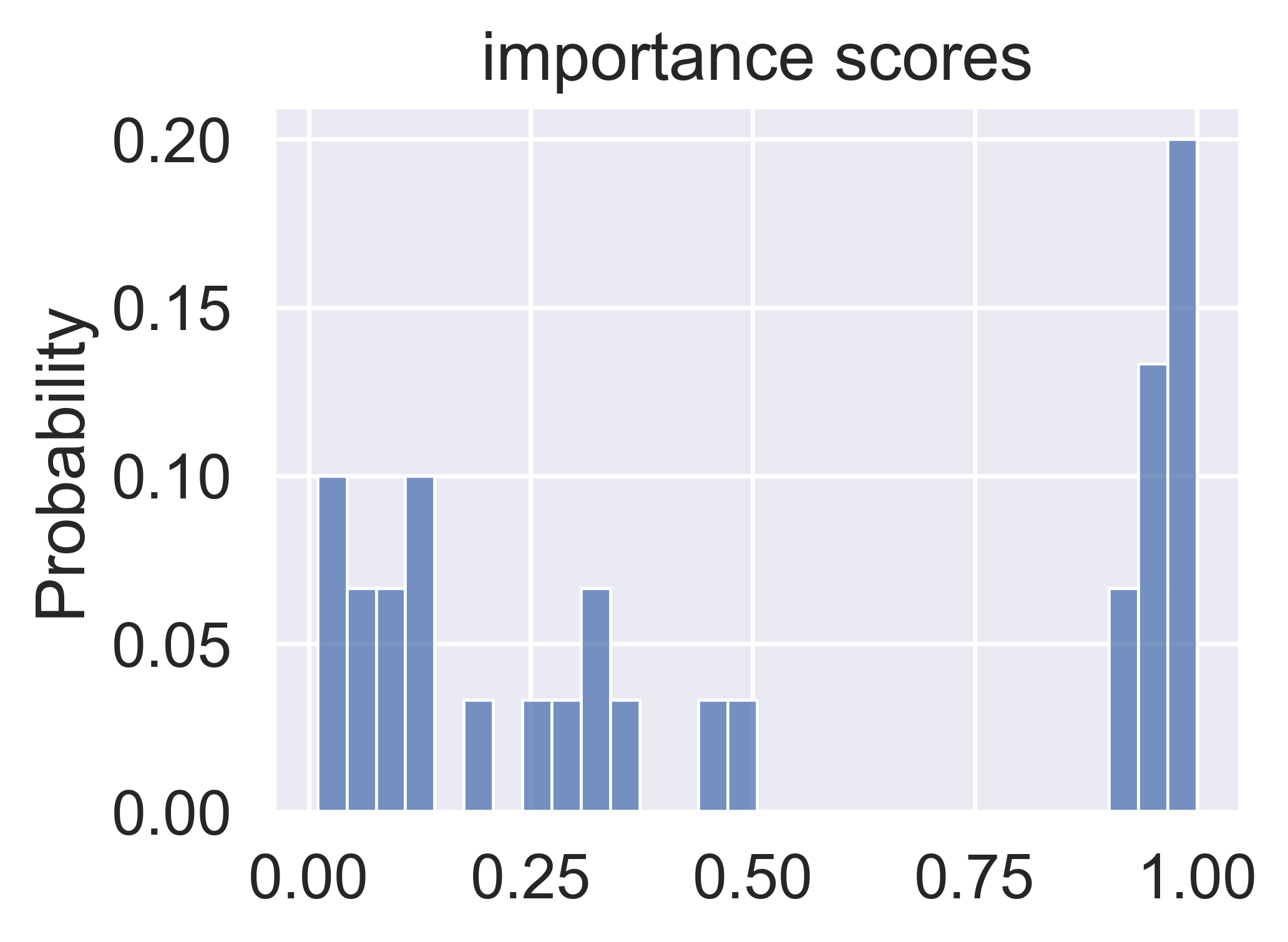}
		\caption{importance score distribution map(trigger size = 0.2)}
		\label{fig:AIDS_comparison}
	\end{figure}

        Figure \ref{fig:AIDS_comparison} shows the map distribution when trigger size is 0.2. Trigger size density is the subgraph’s number of nodes.

	\begin{figure}[h]
		\centering
		\includegraphics[width=0.8\linewidth]{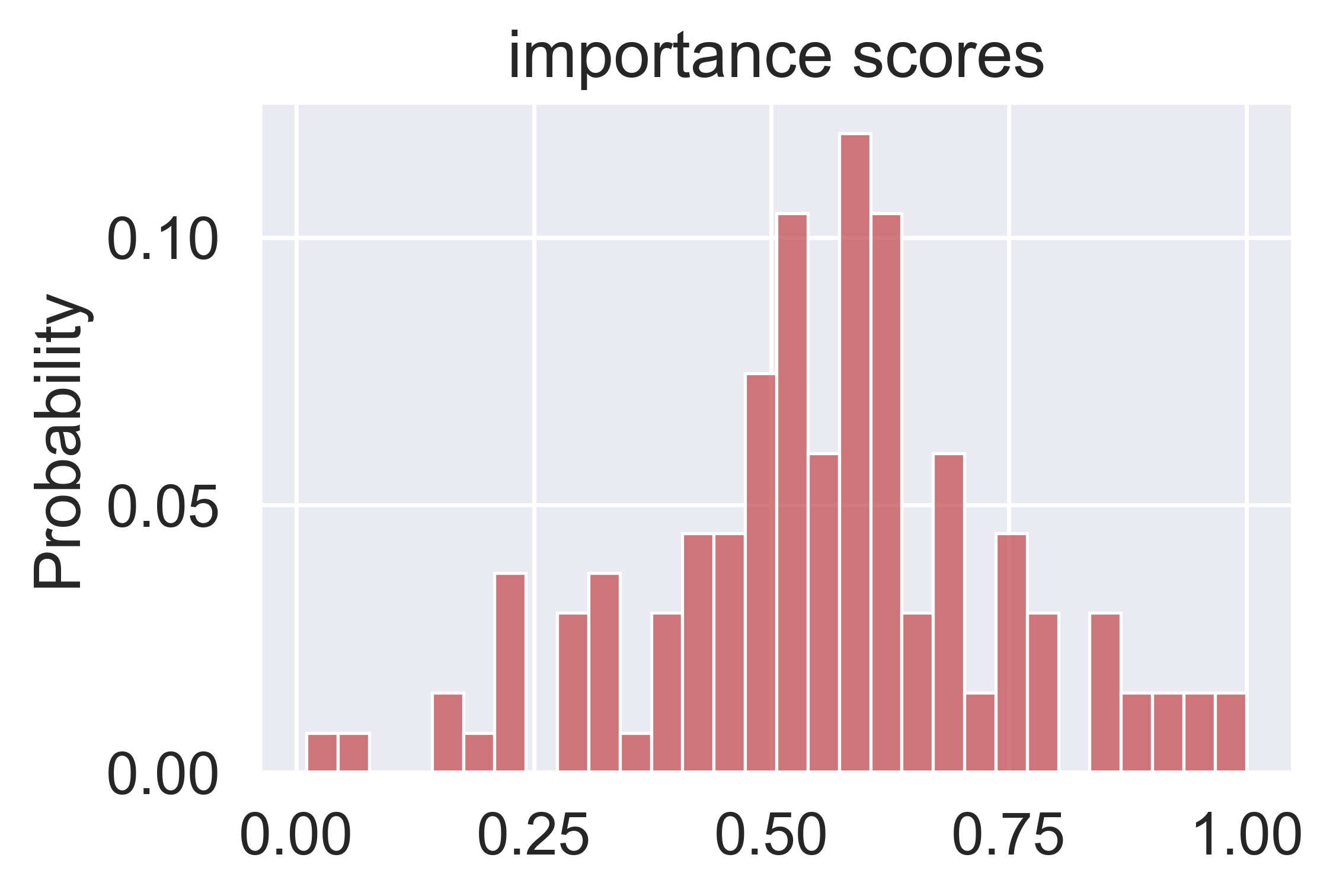}
		\caption{importance score distribution map(trigger size = 0.6)}
		\label{fig:AIDS_comparison2}
	\end{figure}

        On the contrary ,from Figure \ref{fig:AIDS_comparison2}, when trigger size is 0.6, the importance map is more concentrative. Compared with using a fixed threshold, using adaptive sparsity can greatly improve the generalization performance of the model. As a defender, we don't know the information of the attacker's trigger size and structure. Using adaptive sparsity can help us choose an appropriate set of edges to delete. When trigger is big, it means that if we want to remove the whole trigger, we need to delete many edges. In other words, we need a small Sparsity. However, when the trigger is small, we only need to delete a few edges. In this case, big Sparsity is what we need.

        So our goal is to set an appropriate threshold to transform the continuous importance map to a hard importance map only including 0 and 1. Specifically, importance values higher than thresholding will be set to 1 otherwise will be set to 0. 0 represents unimportant edges while 1 represents important edges. After that, we can remove the trigger in the backdoor sample by deleting the edges whose hard importance score is 1. If the trigger size is big, we will need to set relatively low thresholding to generate more "important edges" so that we can delete more edges belonging to the trigger. 

        Until now we set up a relation among trigger size, degree of dispersion of the importance map, and thresholding. We use one evaluation metric called Sparsity to relate them. 

        \subsubsection{Sparsity}, To analyze the performance of explanation methods from the perspective of input graph data, explanation methods should be sparse. it should capture the most important input features and ignore irrelevant features, which can be measured by the Sparsity metric such a feature. Specifically, it measures the score selected as an important feature by the explainability method. Formally, giving the graph $G_i$ and its hard importance map $m_i$, the Sparsity metric can be computed as:

        \begin{equation}
                \text { Sparsity }=\frac{1}{N} \sum_{i=1}^{N}\left(1-\frac{\left|m_{i}\right|}{\left|M_{i}\right|}\right)
        \end{equation}

        where $|m_i|$ denotes the number of important input features (node/edge/node features) identified in $m_i$, and $|M_i|$ denotes the total number of features in the original graph $G_i$. Higher Sparsity values indicate the sparser the model features and tend to only capture the most important input information. Here we use this metric in reverse, that is, we use it to transform the continuous importance map to a hard importance map $m_i$. Actually, sparsity is another form of thresholding. Setting high thresholding equals setting high sparsity. 
        
        Then we introduce the Coefficient of Variation to quantify the degree of dispersion of the importance map:
        
        \begin{equation}
                c_{v}=\frac{\sigma}{\mu}
        \end{equation}   

        where $\sigma$ denotes the standard deviation, $\mu$ denotes the average value. Big $c_v$ represents that the edges' important values are highly discrete. On contrary, small $c_v$ means all the edges' significance is similar. Actually, before the process of calculating $Fidelity$ and $Infidelity$ we firstly set the sparsity via calculate the $c_v$ of the importance map. Then the continuous importance map can be transformed to a hard importance map $m_i$ according to the sparsity and $m_i$ can be used to calculate $Fidelity$ and $Infidelity$.

\begin{figure}[h]
            \centering
            \includegraphics[width=0.7\linewidth]{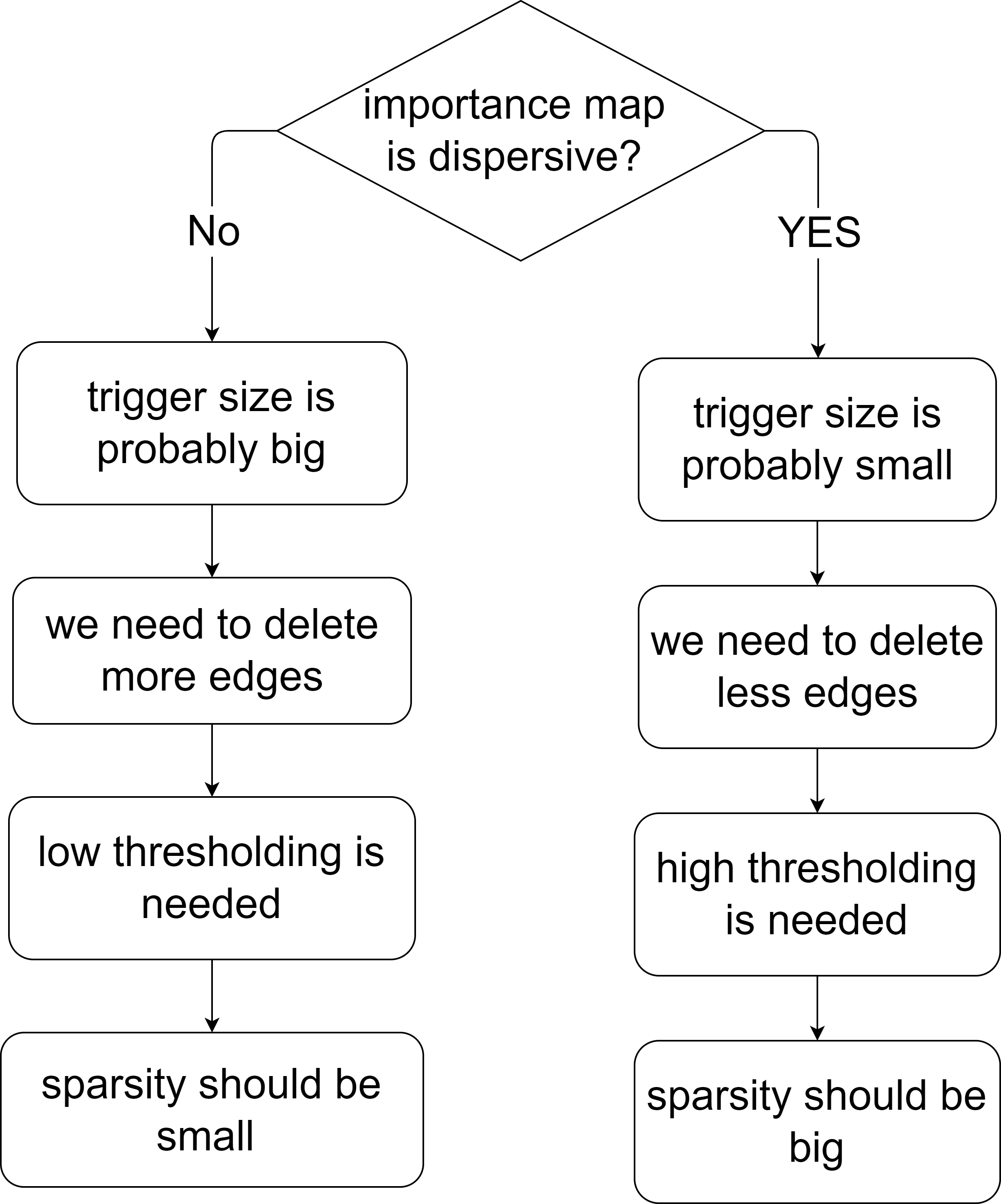}
            \caption{how to deal with different trigger size?}
            \label{fig:flowchar}
        \end{figure}

        Finally, we can put them all together, our reasoning can be summarized in Figure \ref{fig:flowchar}. It solves the problem that how to regulate sparsity. We find that $c_v$ and sparsity have the same trend. So we control the Coefficient of Variation between 0 and 1 to replace the sparsity.

        \subsubsection{Explainability score(ES):},
        To do binary classification of input samples, we introduce a new metric, Explainability score ($ES$). Formally, the can be computed as:

        \begin{equation}
                ES=Fidelity-Infidelity
        \end{equation}   
        
        As we deduced above, backdoor samples always have high fidelity and low infidelity. $ES$ combine both of them together. So high $ES$ value represents a high probability of backdoor sample. In conclusion, there are significant differences in $ES$ between clean and backdoor samples. This provides the defender with a tool for backdoor detection. We use the highest $ES$ value of the validation dataset as the detection boundary. For every incoming input graph, if its $ES$ value is higher than the detection boundary, we will suspect that it's a backdoor sample. In order to disable the backdoor, we delete some most important edges in this input graph according to the hard importance map $m_i$.

	       \begin{figure}[h]
		      \centering
		      \includegraphics[width=0.8\linewidth]{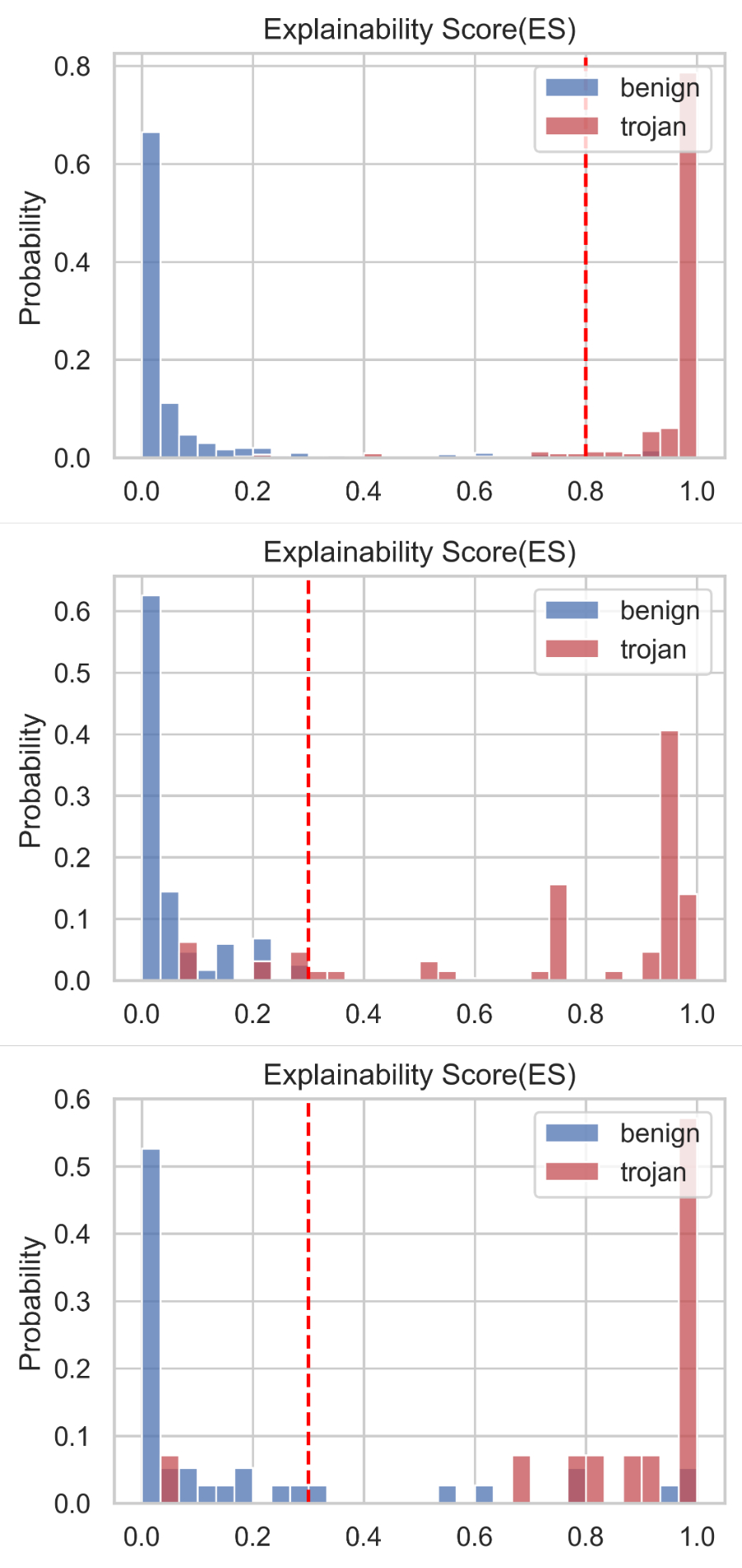}
		      \caption{The differences in ES between clean and contaminated samples}
		      \label{fig:comparison}
	       \end{figure}

        \begin{figure*}[htbp]
        \begin{center}
        \includegraphics[width=0.8\textwidth]{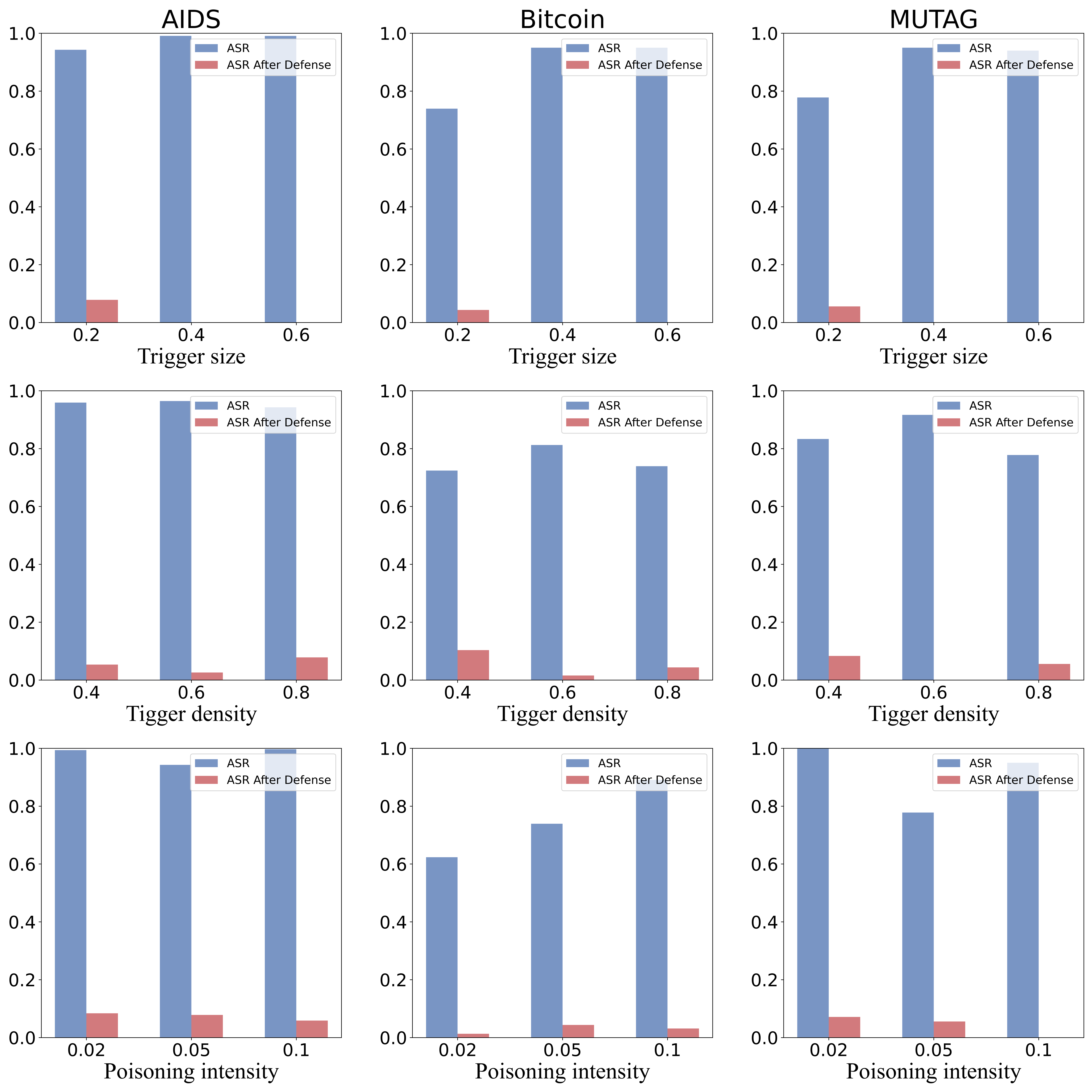}
        \end{center}
        \caption{The effect of our method on ASR indicators}
        \label{fig:ASR_ASR_D}
        \end{figure*}

        \begin{figure*}[htbp]
        \begin{center}
        \includegraphics[width=0.9\textwidth]{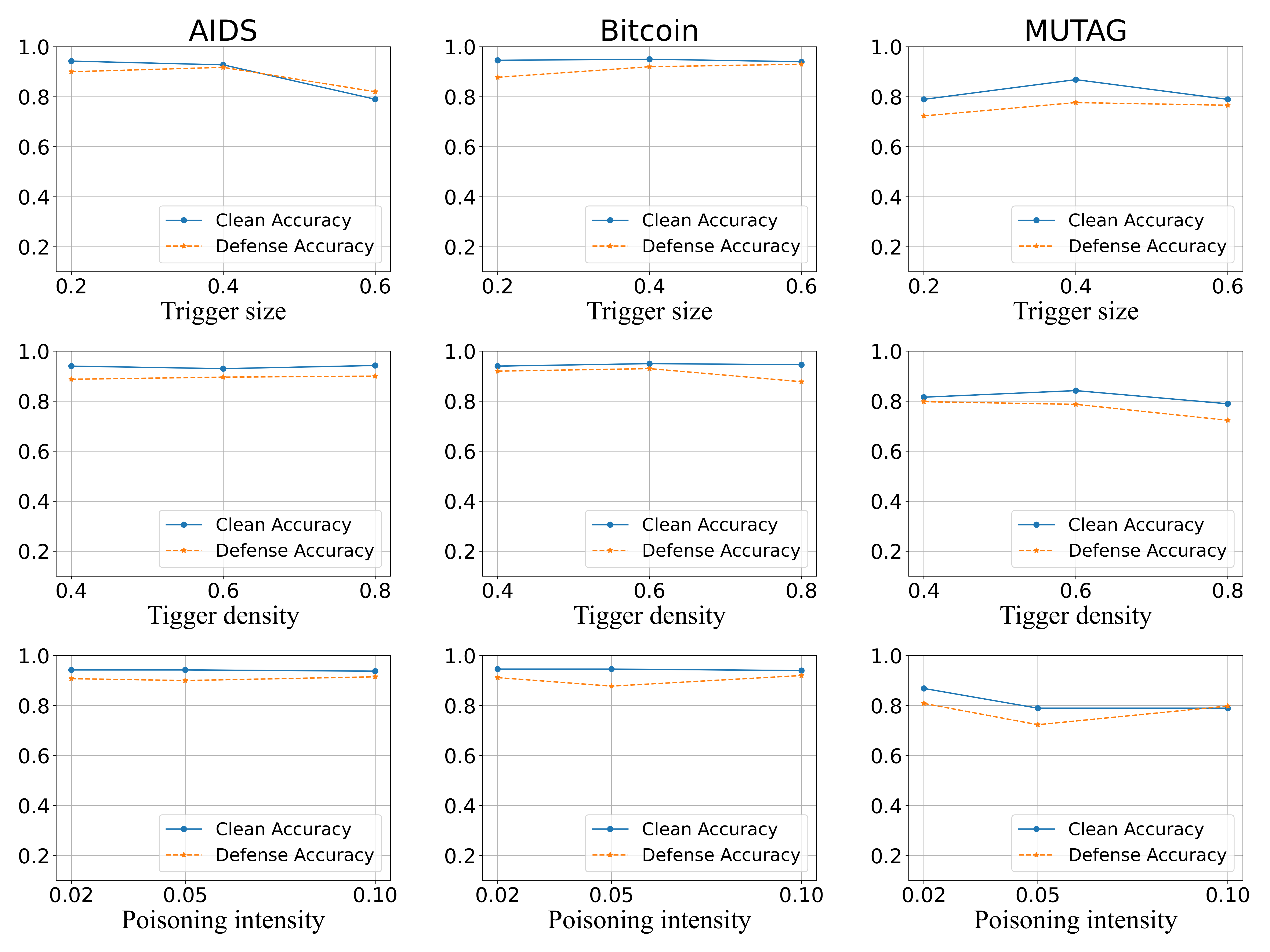}
        \end{center}
        \caption{The effect of our method on ASR indicators}
        \label{fig:clean_defense}
        \end{figure*}

\section{Defense Evaluation}
    \subsection{Experimental settings}
        \subsubsection{Dataset}
            Datasets: We also evaluate our defense on the three datasets
                \begin{itemize}
                    \item[$\bullet$] AIDS \cite{ref47}- molecular structure graphs of active and inactive compounds;
                    \item[$\bullet$] bitcoin \cite{ref14}- an anonymized Bitcoin transaction network with each node (transaction) labeled as legitimate or illicit
                    \item[$\bullet$] MUTAG \cite{ref15}- A dataset of chemical molecules and compounds, with atoms representing nodes and bonds representing edges.
                \end{itemize}

        \subsubsection{explainer:captum} 
        ~\\
        A Pytorch-based model interpretation library. The library provides explainability for many new algorithms (eg: ResNet, BERT, some semantic segmentation networks, etc.) to better understand the specific features, neurons, and neural network layers that contribute to model predictions. For the graph classification problem, it can quickly locate some nodes that affect the results and display them visually.
        
        Integrated gradients is one of the primary attribution algorithm implemented in captum. Integrated gradients is a simple, yet powerful axiomatic attribution method \cite{ref61}. Integrated gradients represent the integral of gradients with respect to inputs along the path from a given baseline to input. Herefore, the integral gradient algorithm only considers the input and output of the model, and the function is differentiable everywhere, without the participation of the internal details of the model(it requires almost no modification of the original network.)
        ~\\

        \subsubsection{models}

        In our evaluation, we use 2 popular GNN models: GraphConv\cite{ref28} and GIN\cite{ref48}to conduct graph classification task. We use their publicly available implementations in $PYG$\cite{ref63}. When a classifier is deployed with our defense system, we denote it as $f_d$.

        ~\\
        \subsubsection{metrics}

       To evaluate attack effectiveness, we use Attack Success Rate $(ASR)$\cite{ref27}, which measures the likelihood that the backdoor classifier predicts the trigger-embedded inputs to the target label  designated by the attacker.

        \begin{equation}
             {Attack \ Success \ Rate}(A S R)=\frac{\# \text { successful trials }}{\# \text { total trials }}
        \end{equation}  		
        
        As for the defender, we want the $ASR$ to drop hugely after we deploy the defense system. Apart from it, our defense system should not influence the model to make a correct prediction on clean samples. In other word, defense system should have a low $False \ Rejection \ Rate(FRR)$ \cite{ref29}. Here we introduce $defense \ accuracy$. Given a clean testing dataset $D_c = \{(G_1,y_1), (G_2,y_2), \cdots, (G_m,y_m) \}$, we define $Defense \ accuracy$ as model's accuracy on clean samples after deploying the defense system. Formally, we have the following:
        
        \begin{equation}
                Defense \ accuracy = \frac{\sum_{i=1}^m \mathbb{I}\left(f_d\left(G_i\right)=y_i\right)}{m}
        \end{equation}    	$f_d$ represents for the model deployed with the defense system. Intuitively, $Defense \ accuracy$ should be close to the clean model's(without deploying defense system) accuracy.

    \subsection{Results}

        Figure \ref{fig:comparison} shows the difference in $ES$ value on three datasets between benign samples and backdoor samples, It turns out that backdoor samples always have extremely high ES. The red dotted line signifies the highest $ES$ in the validation dataset. For every incoming sample, if its $ES$ is higher than it, we will suspect that it's a backdoor sample.

        Because the defender does not know the attacker's trigger information, this disadvantage imposes a requirement on the generalization performance of the defense model. We tested the defense ability under different trigger attributes (Trigger size, Tigger density, Poisoning intensity) and different datasets.

        As shown in Figure \ref{fig:ASR_ASR_D}, in different cases, the $ASR$ of the model decreases significantly after deploying the defense system, For instance, when the trigger size is 20\% (Tigger density=0.8, Poisoning intensity=0.05) of the average number of nodes per graph, The $ASR$ indicator is 0.77 before defense, but the ASR after defense reduces to 0.05 on MUTAG dataset. In some cases, we can even decrease the $ASR$ to 0, which proves the effectiveness of our defense system.

        Figure \ref{fig:clean_defense} shows the result of our defense system on Defense accuracy indicators. it proves that our backdoor defense has a small impact on the accuracies for clean testing graphs. Recall that one of our defender's goals is that the accuracy on clean samples should not be influenced by our defense method. Specifically, Defense accuracy is slightly smaller than clean accuracy. For instance, when the trigger size is 20\% of the average number of nodes per graph, the defense accuracy is 0.04 smaller than the clean accuracy on AIDS dataset. 

        In fact, the importance of features of clean samples is more dispersed than that of malicious samples. The decision result is often determined by the joint action of multiple edges. Even if we misjudge a benign sample to a malicious sample and then we remove some edges from this graph, we can still get the correct decision results from the remaining features. For example, when we distinguish cats and dogs, we judge their categories from multiple features, rather than one feature. Therefore, we seldom make mistakes because of the deletion of a certain feature (such as covering the cat's ears or nose, etc.). On the contrary, as for backdoor sample, the model's decision is almost totally determined by the trigger. Once we remove the trigger, the model's prediction will change a lot. It also proves why backdoor samples will have high $Fidelity$. These properties can reduce accidental damage to clean samples by defensive methods.

\section{Discussion and Limitations}
    First, during the experiment, we found that the model's complexity has a strong correlation with the $ASR$ of backdoor attacks. In the case of low model complexity, when the model achieves the expected accuracy on benign samples, the $ASR$ metric is still at a low level. Therefore, the user is likely to use early stopping\cite{ref62} to avoid overfitting, which indirectly leads to the failure of the attacker's attack. Lower model complexity means that the model's expressive power is weak. Therefore, it is not easy for the model to discover the "importance" of the trigger at a fewer number of training times. Once model ignores the trigger, the backdoor attack will not succeed. On the other hand, when the model's expressive ability is stronger, $ASR$  rises much faster than before. The model can quickly capture the "importance" of the trigger to the classification results, making it easier for attackers to achieve their goals.

	\begin{figure}[h]
		\centering
		\includegraphics[width=0.8\linewidth]{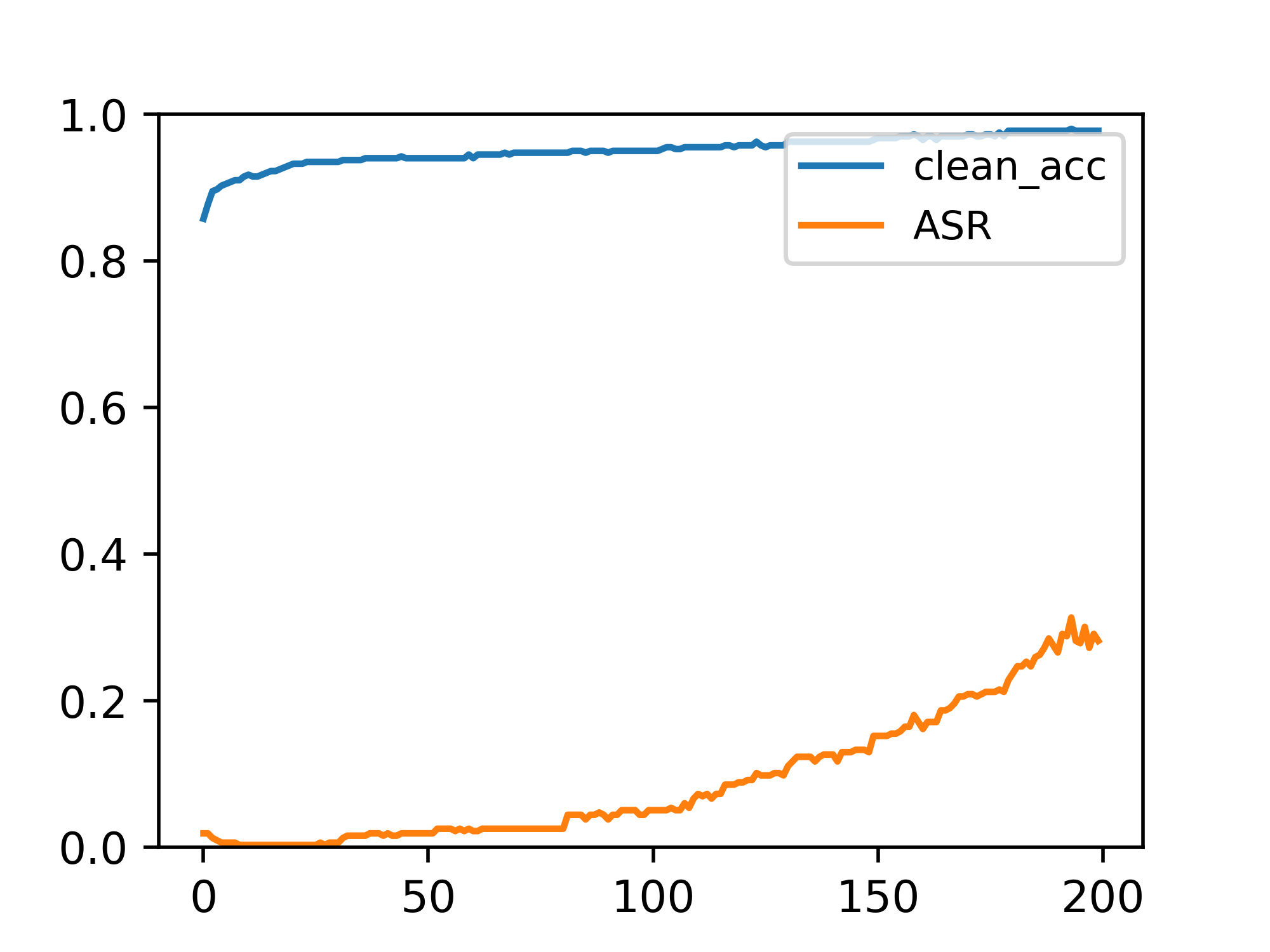}
		\caption{impact of expressive ability of the model (the number of model parameters is 4611)}
		\label{fig:train_chart_fail}
	\end{figure}
        
    As shown in figure \ref{fig:train_chart_fail}, when the expressive ability of the model is insufficient (the number of model parameters is 4611), the $ASR$ rises slowly as the number of epochs increases. And when the accuracy on clean samples reaches the highest value, $ASR$ is still at a low level.  If the victim uses the skill of early stopping and finishes training at epoch=50, it will almost completely invalidate the backdoor attack. From this point of view, backdoor attack is somewhat similar to overfitting.  In essence, overfitting is a phenomenon that the model learns too much about the nonessential features but ignores the essential features in the sample. When overfitting occurs, the model will have an abnormally high training accuracy and a low testing accuracy. Compared with overfitting, the nonessential feature in backdoor attack is the trigger's feature, and the low testing accuracy mentioned above is actually manifested as the model's pretty low accuracy on trigger-embedded samples(They are all misclassified to the target label). At the same time, those triggers also have a pre-defined structure instead of a random structure generated from the nature, which makes the model learns a "regular" noise.  Learning this noise is so easy that the model really trusts it, i.e. the model is overfitting to the trigger.

	\begin{figure}[h]
		\centering
		\includegraphics[width=0.8\linewidth]{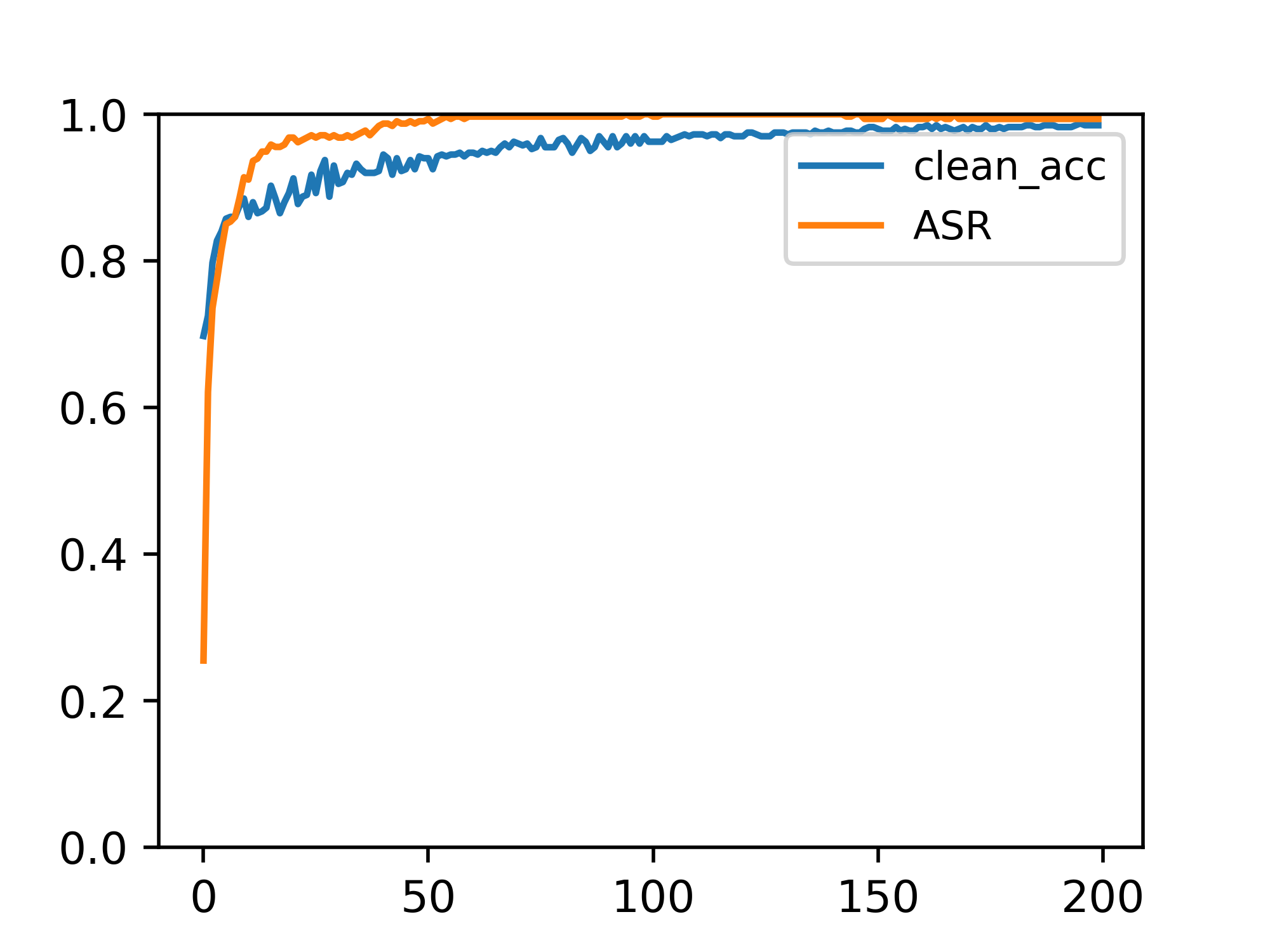}
		\caption{impact of expressive ability of the model (the number of model parameters is 27973)}
		\label{fig:train_chart_success}
	\end{figure}
    
    On the contrary, in figure \ref{fig:train_chart_success}. When the model has the sufficient expressive ability (the number of model parameters is 27973), the $ASR$ rises rapidly and reaches the highest value almost simultaneously with the $clean_acc$ indicator.

	\begin{figure}[h]
		\centering
		\includegraphics[width=0.8\linewidth]{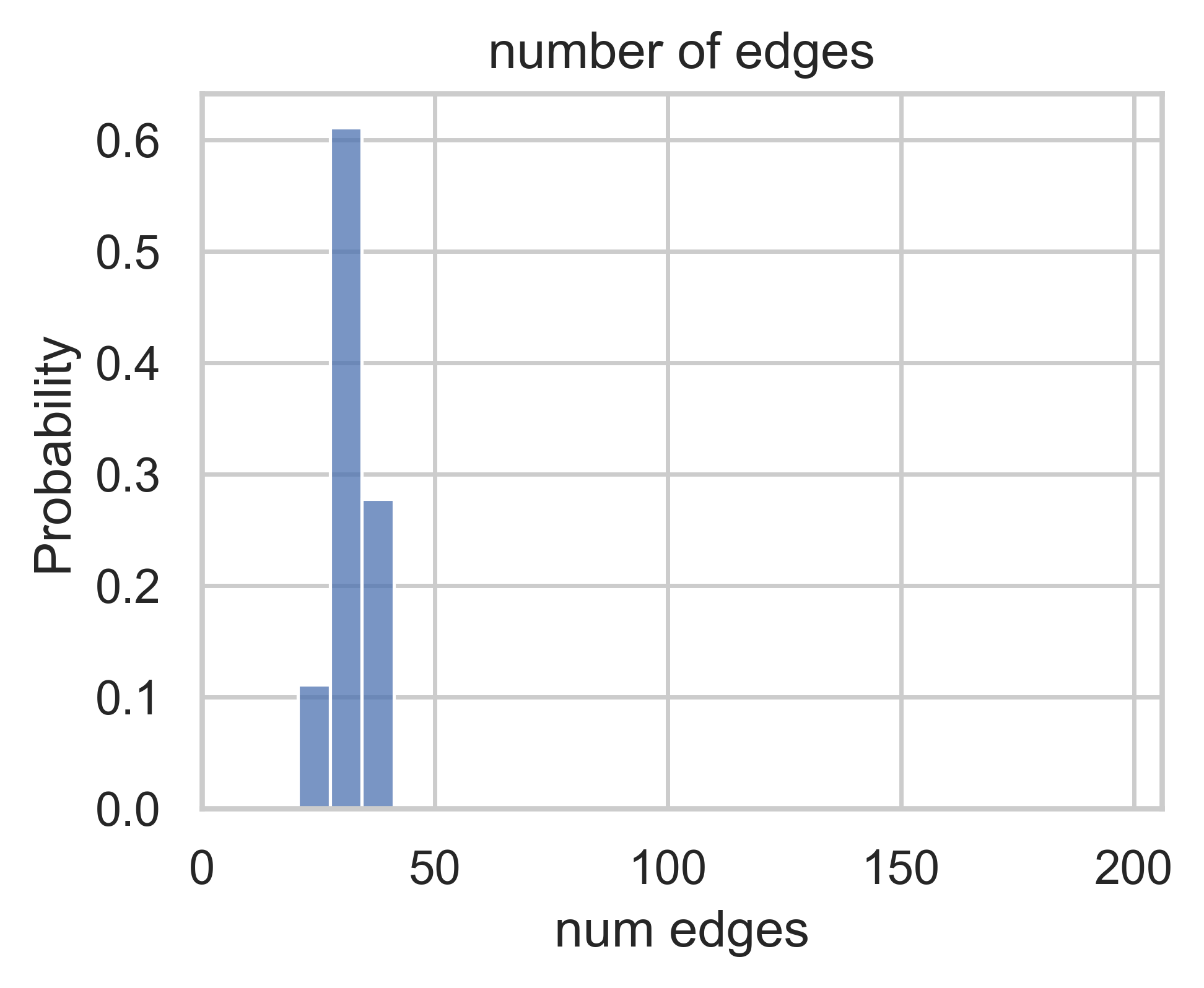}
		\caption{The distribution of undetected malicious samples}
		\label{fig:missed_detection}
	\end{figure}
 
	We also prove that a larger trigger can accelerate the rise of $ASR$.  However, if the trigger is too large, it is easier for users to find errors in the data, so the attacker needs to make a trade-off between the success rate of the attack and the concealment of the trigger.

    We believe that the limitations of this attack algorithm may provide ideas for future research on new defense algorithms (such as using some specific method to make the model's expressive ability	only meet clean samples but unable to learn trigger features)

     The proposed defense method mainly uses the topology information of the graph to judge whether it is a clean graph or a malicious graph. Therefore, there are certain requirements for the topological characteristics of the graph. Specifically, when the input graph is sparse,  each edge is important to the prediction result. The operation of deleting edges is very damaging to the information of the graph. It's hard to tell if its edge is a trigger (or if the graph itself is a trigger)

    Our empirical results in Figure \ref{fig:missed_detection} show that when our defense detection fails to identify the backdoor trigger, the case where the input graph is sparse accounts for the vast majority of the proportion. It shows our method's weakness in discriminating backdoor samples with small size.

\section{Conclusion and Future work}
        In this work, we propose the first backdoor detection and defense method on GNN. Specifically, the defender can calculate the highest ES value in the validation dataset as a threshold to identify the incoming backdoor samples. The samples whose ES is greater than the threshold in the input are regarded as malicious samples. Then remove the edges in it under the control of an adaptive Sparsity so as to achieve the purpose of destroying the trigger. Our empirical evaluation results on three real-world datasets show that our backdoor defense achieves high effectiveness with a small impact on the GNN’s accuracy for clean testing graphs. 

        The explainability and robustness of GNN model is an important issue, and this work provides necessary insights for deeper research. In the future, we are planning to further improve the defense capability of the method on sparse graphs. With the development of the model's explainability, our method is expected to be applied to more complicated engineering problems to further prove its potential for real-life applications.

\bibliographystyle{plain}
 
\bibliography{sample}

\end{document}